%% file: main.tex
\documentclass[11pt, a4paper, logo, two column]{craftjarvis}

\pdftrailerid{redacted}

\makeatletter
\renewcommand\bibentry[1]{\nocite{#1}{\frenchspacing\@nameuse{BR@r@#1\@extra@b@citeb}}}
\makeatother

\input{config}

\usepackage[authoryear, sort&compress, round]{natbib} 

\title{\includegraphics[scale=0.15,valign=c]{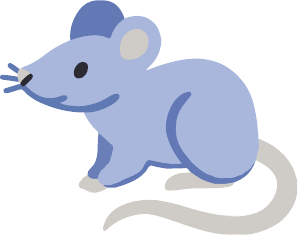}~RAT: Retrieval Augmented Thoughts Elicit Context-Aware Reasoning in Long-Horizon Generation}

\reportnumber{} 

\renewcommand{\today}{March 2024} 

\author[1]{Zihao~Wang}
\author[2]{Anji~Liu}
\author[1]{Haowei~Lin}
\author[3]{Jiaqi~Li}
\author[3]{Xiaojian~Ma}
\author[1]{Yitao~Liang}

\correspondingauthor{
Yitao~Liang~\texttt{yitaol@pku.edu.cn}\\
<Zihao~Wang>~\texttt{zhwang@stu.pku.edu.cn}~<Anji~Liu>~\texttt{liuanji@cs.ucla.edu}
<Haowei~Lin>~\texttt{linhaowei@pku.edu.cn}~\\<Jiaqi Li>~\texttt{ljqjane@gmail.com}~<Xiaojian~Ma>~\texttt{xiaojian.ma@ucla.edu}
}

\affil[1]{Peking~University}
\affil[2]{University~of~California,~Los Angeles}
\affil[3]{Beijing~Institute~for~General~Artificial~Intelligence}

\begin{abstract}
\input{abstract.tex}

\end{abstract}

\begin{document}


\maketitle

\input{sections}


\section*{Limitations}

In this seciton, we discuss three limitations of our \method as follows.

One limitation of this work is that the performance of \method relies on the chain-of-thought reasoning and in-context learning (or RAG) capability of the base LLM. Since this work does not involve any model training, the capability of base LLM will not change when applying \method. Despite \method achieves significant improvement on powerful LLMs such as GPT-3.5 and GPT-4, the effect on smaller and weaker LLMs such as GPT-2 is questionable. On top of that, it is interesting to further explore how to improve \method via fine-tuning weaker LLMs~\citep{ke2023continual,lin2024selecting}.

Another limitation of this work is that the performance of \method also relies on the quality of the retrieved knowledge. When we have an inferior external knowledge base which is irrelevant to the user query, the retrieved knowledge may be unhelpful for LLMs to generate useful information. Also, even if we select a relatively large knowledge base that entails the relevant information, it will be expensive to maintain and retrieve from such a huge knowledge base and also hurts the retrieval precision. An interesting and crucial direction is to study how to build and evaluate the quality of a knowledge base used for efficient and effective retrieval.

It is noteworthy that the above two limitations also apply to the traditional studies on retrieval-augmented generation (RAG). The last limitation of \method is that we follow CoT to solve the problems in a explicit step-by-step fashion. Sometimes step-by-step thinking may be redundant for straightforward questions, while some questions require more complex reasoning structures (e.g., tree-of-thoughts~\citep{treeofthought}). It is also interesting to explore the better reasoning methods for LLMs in our future work.

\section*{Ethics Statement}

All datasets and models are publicly accessible except for OpenAI's GPT series and the text embedding APIs. We have not identified any significant ethical considerations associated with this work. We believe our newly proposed \method can improve the generation of LLMs in various fields and reduce LLMs' hallucinations.

\bibliographystyle{abbrvnat}
\nobibliography*
\bibliography{main}

\newpage

\appendix
\renewcommand\thefigure{\thesection.\arabic{figure}}
\setcounter{figure}{0}
\input{appendix}

\end{document}

%% file: config.tex
\usepackage{xcolor}
\definecolor{Gray}{gray}{0.9}
\definecolor{mygreen}{rgb}{0.0, 0.5, 0.0}
\definecolor{myred}{rgb}{0.8, 0.25, 0.33}
\definecolor{myblue}{rgb}{0.19, 0.55, 0.91}
\definecolor{uclablue}{rgb}{0.15, 0.45, 0.68}
\definecolor{zihaoblue}{rgb}{0.25, 0.44, 0.88}
\definecolor{zihaored}{rgb}{0.79, 0.36, 0.27}
\definecolor{boxgreen}{rgb}{0.02, 0.66, 0.02}
\definecolor{boxred}{rgb}{0.66, 0.1, 0.1}
\definecolor{boxblue}{rgb}{0.01, 0.01, 0.73}

\definecolor{mygray}{gray}{0.4}

\usepackage{times}
\usepackage{graphicx}
\usepackage{amssymb}
\usepackage{url}

\usepackage{microtype}
\usepackage{subfigure}
\usepackage{booktabs}

\usepackage{amsmath}
\usepackage{mathtools}
\usepackage{amsthm}

\usepackage{listings}
\usepackage{etoc}
\usepackage{algorithm}
\usepackage{lipsum}
\usepackage{pifont}
\usepackage{wrapfig}
\usepackage{colortbl}
\usepackage{acronym}
\usepackage{multicol}
\usepackage{multirow}
\usepackage{makecell}
\usepackage{enumitem}
\usepackage{tabularx}
\usepackage{colortbl}
\usepackage{array}
\usepackage{xspace}
\usepackage[skip=3pt,font=small]{caption}
\usepackage{xspace}
\usepackage{mdframed}

\usepackage[export]{adjustbox}

\usepackage[utf8]{inputenc} 
\usepackage[T1]{fontenc}    
\usepackage{amsfonts}       
\usepackage{nicefrac}       

\usepackage{algpseudocode}

\usepackage{mathtools}
\usepackage{amssymb}
\usepackage{amsthm}

\renewcommand{\paragraph}[1]{\noindent\textbf{#1.}}

\makeatletter
\DeclareRobustCommand\onedot{\futurelet\@let@token\@onedot}
\def\@onedot{\ifx\@let@token.\else.\null\fi\xspace}
\def\eg{\emph{e.g}\onedot} 

\def\ie{\emph{i.e}\onedot}

\def\etc{\emph{etc}\onedot}

\makeatother

\acrodef{llms}[LLMs]{Large Language Models}
\acrodef{mlms}[MLMs]{Multimodal Language Models}
\acrodef{rag}[RAG]{Retrieval Augmented Generation}
\acrodef{cot}[CoT]{chain-of-thought}
\acrodef{rat}[RAT]{Retrieval Augmented Thoughts}
\newcommand{\method}{\ac{rat}\xspace}

\usepackage{xcolor}

\usepackage[most]{tcolorbox}

\newtcolorbox[list inside=prompt,auto counter,number within=section]{prompt}[1][]{
    colbacktitle=black!60,
    coltitle=white,
    fontupper=\footnotesize,
    boxsep=5pt,
    left=0pt,
    right=0pt,
    top=0pt,
    bottom=0pt,
    boxrule=1pt,
    #1,
}

\newtcolorbox[auto counter,number within=section]{codebox}[2][]{
    listing only,
    title={Code Example \thetcbcounter: #2},
    #1,
}

\newtcolorbox[list inside=response,auto counter,number within=section]{response}[1][]{
    colbacktitle=black!40,
    coltitle=white,
    fontupper=\footnotesize,
    boxsep=5pt,
    left=0pt,
    right=0pt,
    top=0pt,
    bottom=0pt,
    boxrule=1pt,
    #1,
}

%% file: abstract.tex
We explore how iterative revising a chain of thoughts with the help of information retrieval significantly improves large language models' reasoning and generation ability in long-horizon generation tasks, while hugely mitigating hallucination. 
In particular, the proposed method --- \emph{retrieval-augmented thoughts} (RAT)
--- revises each thought step one by one with retrieved information relevant to the task query, the current and the past thought steps, after the initial zero-shot CoT is generated.
Applying RAT to GPT-3.5, GPT-4, and CodeLLaMA-7b substantially improves their performances on various long-horizon generation tasks; on average of relatively increasing rating scores by 13.63\% on code generation, 16.96\% on mathematical reasoning, 19.2\% on creative writing, and 42.78\% on embodied task planning. 

The demo page can be found in \url{https://craftjarvis.github.io/RAT}.

%% file: sections.tex
\begin{figure*}[t]
    \centering
    \includegraphics[width=\linewidth]{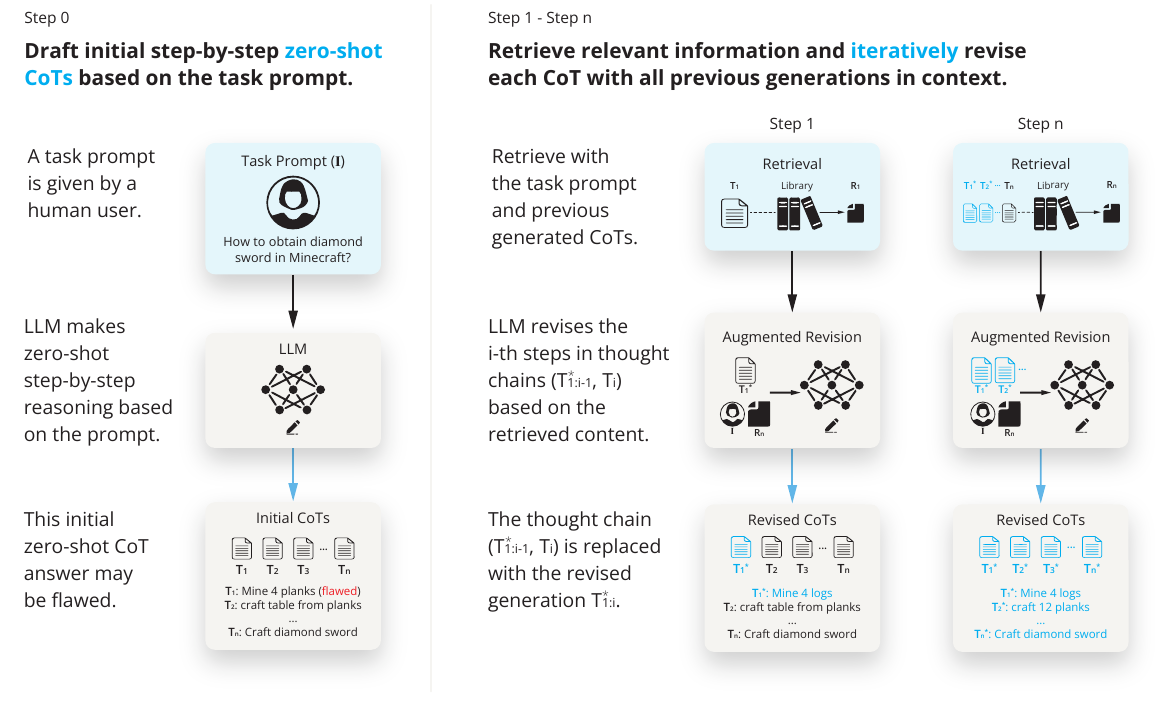}
    \caption{
        \textbf{Pipeline of \method.} Given a task prompt (denoted as $\mathit{I}$ in the figure), \method starts from initial step-by-step thoughts ($T_1,T_2,\cdots,T_n$) produced by an LLM in zero-shot (``let's think step by step''). Some thought steps (such as $T_1$ in the figure) may be {\color{red}flawed} due to hallucination. \method iteratively revises each thought step $(T^\star_1, T^\star_2, \cdots, T^\star_{i-1}, T_i)$ using RAG from an external knowledge base (denoted as \texttt{Library}). Detailed prompting strategy can be found in \autoref{sec:rat_method}.
    }
    \label{fig:pipeline}
    \vspace{-0.2 in}
\end{figure*}

\begin{figure*}[t]
    \centering
    \includegraphics[width=\linewidth]{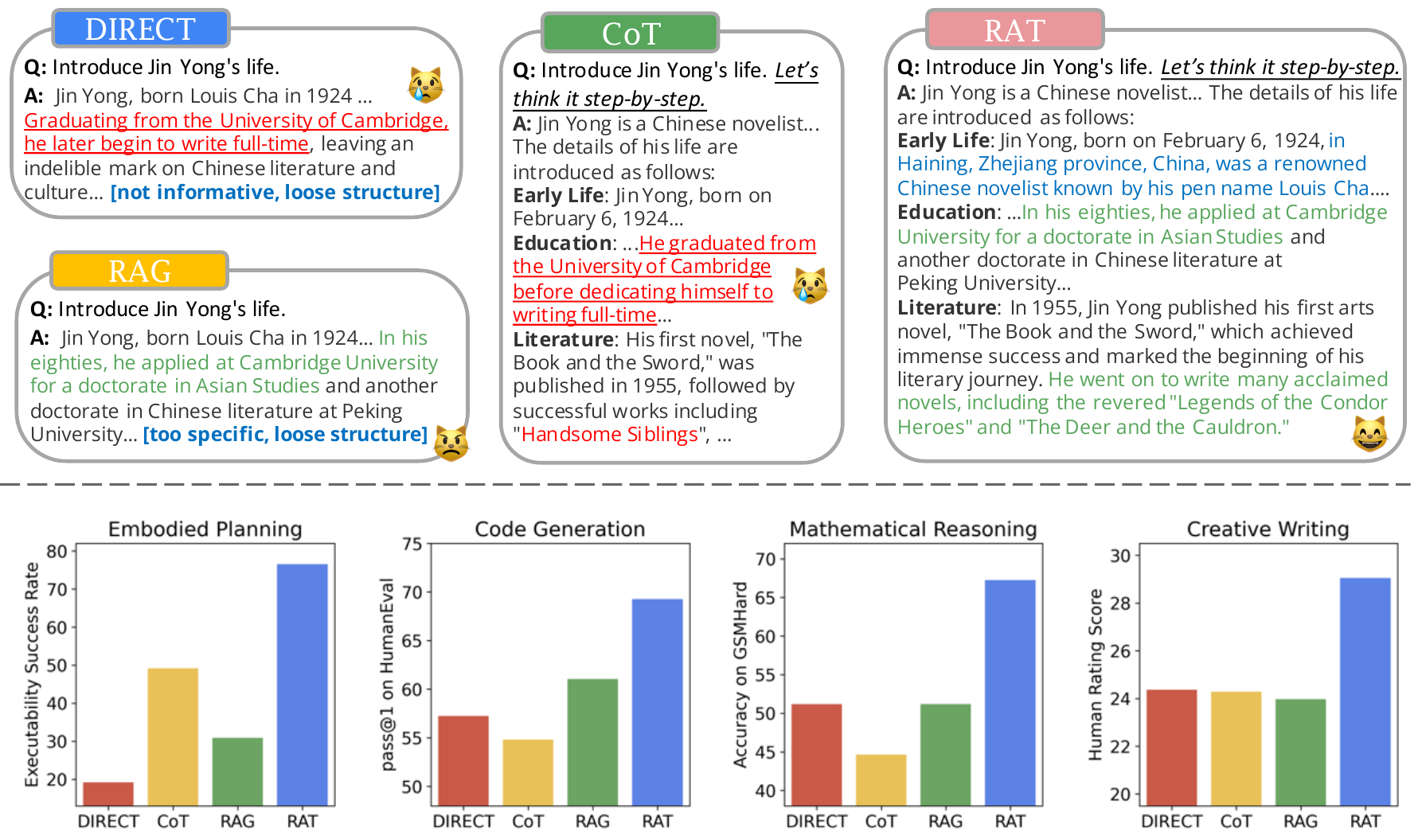}
    \caption{
        \textbf{Top}: An example of different LLM reasoning methods on creative generation tasks. 
        \textcolor{red}{Red} text indicates errors or illusions in the text generated by LLM, while \textcolor{mygreen}{green} text represents correct generation. Methods without RAG often generate incorrect information with hallucination, classical RAG is highly related to retrieved content with a loose structure, and RAT-generated texts perform best in terms of accuracy and completeness.
        \textbf{Bottom}: The quantitative performance comparison for different LLM reasoning methods on complex embodied planning, mathematical reasoning, code generation, and creative generation tasks. Our \method outperforms all the baselines on all tasks.
    }
    \label{fig:overall_results}
    \vspace{-0.2 in}
\end{figure*}

\section{Introduction}
\ac{llms} have achieved fruitful progress on various natural language reasoning tasks~\citep{chainofthought,react,self-consistency,least-to-most,gpt3}, especially when combining large-scale models ~\citep{chowdhery2022palm,gpt4} with sophisticated prompting strategies, notably \ac{cot} prompting~\citep{chainofthought,kojima2022large}. However, there have been increasing concerns about the factual correctness of \ac{llms} reasoning, citing the possible hallucinations in model responses~\citep{hallucination} or the intermediate reasoning paths, \ie \ac{cot}s~\citep{dhuliawala2023chainofverification}. This issue becomes more significant when it comes to zero-shot \ac{cot} prompting, aka. ``let's think step-by-step''~\citep{kojima2022large} and long-horizon generation tasks that require multi-step and context-aware reasoning, including code generation, task planning, mathematical reasoning, \etc. Factually valid intermediate thoughts could be critical to the successful completion of these tasks. 

Several prompting techniques have been proposed to mitigate this issue, one promising direction, \ac{rag}~\citep{rag} seeks insights from human reasoning~\citep{holyoak2012oxford}, and utilizes retrieved information to facilitate more factually grounded reasoning.
In this paper, we explore how to synergize \ac{rag} with sophisticated long-horizon reasoning. Our intuition is that the hallucination within the intermediate reasoning process could be alleviated through the help of outside knowledge. The resulting prompting strategy, \textit{retrieval-augmented thoughts} (RAT), is illustrated in \autoref{fig:pipeline}. Our strategy comprises two key ideas. Firstly, the initial zero-shot \ac{cot} produced by \ac{llms} along with the original task prompt will be used as queries to retrieve the information that could help revise the possibly flawed \ac{cot}. Secondly, instead of retrieving and revising with the full \ac{cot} and producing the final response at once, we devise a progressive approach, where \ac{llms} produce the response step-by-step following the \ac{cot} (a series of subtasks), and only the current thought step will be revised based on the information retrieved with task prompt, the current and the past CoTs. This strategy can be an analogy to the human reasoning process: we utilize outside knowledge to adjust our step-by-step thinking during complex long-horizon problem-solving~\citep{holyoak2012oxford}. A comparison of \method and counterparts can be found in \autoref{fig:overall_results}.

We evaluate \method on a wide collection of challenging long-horizon tasks, including code generation, mathematical reasoning, embodied task planning, and creative writing. We employ several \ac{llms} of varied scales: GPT-3.5~\citep{gpt3}, GPT-4~\citep{gpt4}, CodeLLaMA-7b~\citep{codellama}. The results indicate that combing \method with these \ac{llms} elicits strong advantages over vanilla \ac{cot} prompting and \ac{rag} approaches. In particular, we observe new state-of-the-art level of performances across our selection of tasks: 1) code generation: HumanEval (+20.94\%), HumanEval+ (+18.89\%), MBPP (+14.83\%), MBPP+ (+1.86\%); 2) mathematical reasoning problems: GSM8K (+8.36\%), and GSMHard (+31.37\%); 3) Minecraft task planning (2.96 times on executability and +51.94\% on plausibility); 4) creative writing (+19.19\% on human score). Our additional ablation studies further confirm the crucial roles played by the two key ingredients of \method: revising \ac{cot} using \ac{rag} and progressive revision \& generation. This work reveals how can \ac{llms} revise their reasoning process in a zero-shot fashion with the help of outside knowledge, just as what humans do.

\section{Retrieval Augmented Thoughts}
Our goal is to support long-horizon reasoning and generation while mitigating hallucination when using LLMs. To have satisfying performance on long-horizon tasks, two ingredients are indispensable. Firstly, access to factual information can be facilitated by retrieval. Secondly, appropriate intermediate steps that outline a scratchpad to finish complex tasks, can be facilitated by CoT. Yet, a naive combination of the two would not necessarily yield improvements. Two questions still persist: (1) what is relevant information to retrieve; (2) how to effectively correct reasoning steps with relevant factual information. To better appreciate our method and why our method can address these two questions, we first provide a brief preliminary introduction of RAG and CoT. 

\begin{algorithm*}
\caption{Retrieval augmented thoughts (RAT)}
\label{alg:pipeline}
\begin{algorithmic}[1]
\Require Task Prompt $I$, Autoregressive Large Language Model $p_\theta$
\State $T=\{T_1, T_2, \ldots, T_n\} \gets p_\theta(\cdot|I)$ \Comment{Generate zero-shot initial step-by-step thoughts $T$}
\State $T^\star \gets T_1, i \gets 1$ \Comment{Draft answer $T^\star$ initialized with the first thought step $T_1$}
\Repeat
    \State $Q_i \gets \text{ToQuery}(I, T^\star)$ \Comment{Generate query $Q_i$ based on current draft answer $T^\star$}
    \State $R_i \gets \text{RetrieveFromCorpus}(Q_i)$ \Comment{Retrieve information $R_i$ from corpus or Internet}
    \State $T^\star \gets p_\theta(\cdot|I, T^\star, R_i)$ \Comment{Revise draft answer $T^\star$ based on retrieved text $R_i$}
    \State $T^\star \gets \text{CONCAT}(T^\star, T_{i+1})$ \Comment{Append the next thought step $T_{i+1}$}
    \State $i \gets i + 1$ \Comment{Begin the next revision round}
\Until{$i > n$} \Comment{Repeat until all the revised thoughts $T_{1:n}^\star$ are obtained}
\State \textbf{return} $T^\star$ \Comment{Output $T^\star$ as the final generation}
\end{algorithmic}
\end{algorithm*}

\subsection{Preliminary}\label{sec:preliminary}

\noindent \textbf{Retrieval-Augmented Generation} (RAG) targets the problem of generating fictitious facts by providing LLMs with relevant text extracted from trusted sources. It is primarily used in question-answering (QA) tasks~\citep{rag}. Specifically, given a set of $n$ candidate documents $R := \{R_{i}\}_{i=1}^{n}$, RAG aims to retrieve the most relevant ones w.r.t. a query $Q$, which can be the question/task prompt itself or relevant information generated by LLMs. To achieve this, RAG first extracts semantic-aware embeddings of the documents $r_i := \text{emb} (R_i) \in \mathbb{R}^{K}$ ($K$ is the size of the embedding) as well as the query $q := \text{emb} (Q) \in \mathbb{R}^{K}$. $\text{emb}(\cdot)$ can be implemented with various text embedding models, such as Sentence-BERT~\citep{sentencebert}. The relevance between the query and a document is measured by their cosine similarity:
\begin{align*}
    \text{sim}(Q, R_i)=\frac{q\cdot r_i}{\|q\|\|r_i\|}.
\end{align*}
Based on their relevance, the top-ranked $k$ documents are then fed into the prompt for LLMs to generate the final answer. With such rich and factual contexts, RAG mitigates the hallucination of LLMs. However, complex reasoning tasks (e.g., those requiring multi-step reasoning) can be difficult to translate into effective search queries, leading to challenges in finding relevant documents and making RAG less applicable. Traditionally, RAG retrieves all relevant information at once. Yet, it overlooks the fact that it is difficult to predict what ``facts" or information is required in the subsequent reasoning and generation steps. The task prompt itself is hardly sufficient to provide enough clues for this.

\noindent \textbf{Chain of Thoughts (CoT) prompting} is designed to enhance the performance of LLMs under tasks that require complex reasoning steps~\citep{chainofthought}, such as multi-step math word problems. Specifically, instead of tasking LLMs to generate the correct answer directly, CoT prompting incentivizes LLMs to first output intermediate reasoning steps, termed thoughts, that serve as a scratch space for the task, before summarizing the thoughts into a final answer. Such behavior of LLMs can either be stimulated in \textit{zero-shot} by prompting terms that encourage CoT reasoning (e.g., ``let's think step by step'')~\citep{kojima2022large}, or triggered by few-shot examples that perform CoT in similar tasks. However, since no direct supervision is posed to the intermediate thoughts, LLMs could make errors due to the lack of relevant domain knowledge~\citep{llama2} or biased by hallucinations~\citep{hallucination}.

\subsection{Our Approach}\label{sec:rat_method}

Our intuition to mitigate the issues of CoT prompting and RAG mentioned above is to apply RAG to revise every thought step generated by CoT prompting. An overview can be found in \autoref{fig:pipeline} and Algorithm~\ref{alg:pipeline}. Specifically, given a task prompt $\mathit{I}$, we first prompt LLM to generate step-by-step thoughts in \textit{zero shot} (``let's think step-by-step'') $T := \{T_i\}_{i=1}^{n}$, where $T_i$ represents the $i$th thought step. In long-horizon generation tasks, $T$ can either be the intermediate reasoning steps, \eg the pseudo code with comments in code generation, article outline in creative writing, \etc, or the draft response itself, \eg a list of sub-goals in embodied task planning as shown in \autoref{fig:pipeline}. 

Since $T$ could be flawed (\eg, contains hallucination), we proceed to use RAG to revise every generated thought step before generating the final response from these thoughts. Specifically, assuming we have fixed the previous thought steps and now are about to revise $T_{1:i}$, we begin by converting the text $\{\mathit{I}, T_1, \dots, T_i\}$ into a query $Q_i$:
\begin{align*}
    Q_i=\text{ToQuery}(\mathit{I},T_1,\dots,T_i),
\end{align*}
\noindent
where $\text{ToQuery}(\cdot)$ can either be a text encoder or an LLM that translates the task prompt $\mathit{I}$, the current and the past thought steps $T_1, \dots, T_i$ into a query $Q_i$ that can be processed by the retrieval system. We adopt RAG to retrieve relevant documents $R_i$ using $Q_i$, which are then prepended to the prompt to generate a revised thought step $T^\star_i$.
\begin{align*}
    T^\star_{1:i} =p_\theta(\cdot|\mathit{I},T_1,\dots,T_i,R_i).
\end{align*}
\noindent
Finally, depending on the actual task, the revised thought steps $T^\star_{1:n}$ can simply be used as the final model response, \eg, embodied task planning. For tasks like code generation, or creative writing, the LLM will be further prompted to produce the complete response (code, passage) from each revised thought step in a step-by-step fashion.

Note that, when revising the $i$-th thought step $T_i$, instead of using the current step $T_i$ only, or the complete chain of thoughts $T_1, \dots, T_n$ to produce the query for RAG, we ensure the query $Q_i$ is produced from the current thought step $T_i$ and previous revised thought steps $T_{1:i-1}^\star$, \ie, we adopt a \textit{casual reasoning} to revise the thoughts using RAG: 

\begin{align*}
    Q_i &= \text{ToQuery}(I, T_{1:i-1}^\star, T_i)\\
    T_{1:i}^\star &= p_\theta(\cdot|I,T_{1:i-1}^\star, T_i, R_i).
\end{align*}

This allows for the correction of errors in the original thoughts $T$ by continually consulting different reference texts and ensures that each step of reasoning is informed by the most accurate and relevant information, significantly improving the quality and reliability of the generated output. 

Our hypothesis why our method can address the two problems mentioned at the beginning of this section is as follows. Firstly, the most straightforward way to know what information will be used in complex reasoning is to ``see'' the reasoning steps. Our approach leverages all the generated thoughts along with the task prompt to provide more clues for more effective retrieval. Secondly, some information cannot be directly retrieved, especially information related to the final answer to a hard complex question. Instead, retrieval of information relevant to intermediate questions, which are assumed to be easier, is more accessible. Thanks to the compositional nature of many reasoning tasks, an iterative retrieval process could also be more effective. Thirdly, correcting potential hallucinations needs to be targeted. Revising a complete CoT with RAG could introduce errors at otherwise already-correct steps. Revising every step one by one could be more reliable. The first two points address question (1) and the last point addresses question (2). Quantitative evidence can be found in our ablation studies in \autoref{sec:exp_ablation}.

\section{Experiments}
We test our proposed method \method on a diverse set of benchmarks that highlight long-horizon generation and reasoning. Existing methods traditionally struggle in those benchmarks; ``hallucinated" steps are obvious in LLMs' outputs. Those steps either fail to stick to the original query or are plainly invalid. We kindly refer readers to subsection 3.3 (case analysis) for a more detailed discussion. Due to space constraints, we do not introduce each benchmark setting, nor do we discuss our results in each benchmark in full length. Rather, this section provides a comprehensive demonstration of our method's performance and provides a spotlight to provide preliminary empirical analysis about why and when our method works and when it fails.

\subsection{Experimental Setups}
We adopt four groups of benchmarks.

\input{tables/code_results}

\input{tables/merge_table}

\noindent \textbf{Code Generation} includess HumanEval~\citep{codex}, HumanEval+~\citep{evalplus}, MBPP~\citep{mbpp}, and MBPP+~\citep{evalplus}. These benchmarks encompass a wide range of programming problems, from simple function implementations to more complex algorithmic challenges, providing a robust testbed for assessing generative capabilities. 
    
\noindent\textbf{Mathematical Reasoning} evaluation is conducted on GSM8K and GSM-HARD dataset, which comprises thousands of multi-step mathematical problems~\citep{gsm8k,pal}.

\noindent \textbf{Creative Writing} tasks are conducted to evaluate the versatility of \method, including survey, summarization \etc., highlighting different aspects of open-ended text generation.

\noindent \textbf{Embodied Planning} tasks are evaluated on open-ended environments Minecraft. A set of 100 tasks ranging from simple objectives to challenging diamond objectives are evaluated through MC-TextWorld~\citep{mcu}. 

\paragraph{Evaluation Metrics} 
For code generation, the classical pass rate pass@k is selected as the evaluation metrics~\citep{codex,evalplus}, $k$ denotes the sampling number. We compute accuracy to evaluate every question in mathematical reasoning tasks, aligning with the established metric for the GSM8K~\citep{gsm8k}. For embodied planning tasks, we compute the plan execution success rate in MC-TextWorld as executability~\citep{mcu}. We also conduct human elo rating evaluation to compute the trueskill rating score~\citep{trueskill} for embodied planning (as plausibility) and creative writing tasks. These indicators are better the higher they are.

\paragraph{Baselines}
To establish a comprehensive and equitable comparison landscape, we incorporate a suite of baseline methods. Our baselines include the original language models, referred to as DIRECT, and the Retrieval-Augmented Generation (RAG) methodology with $n$ retrieved examples, instantiated in both single-shot (1 shot) and multi-shot (5 shots) configurations, as documented by \citet{rag}. Additionally, we examine the zero-shot \ac{cot} (CoT) approach, as conceptualized by \citet{kojima2022large}, which simulates a step-by-step reasoning process to facilitate complex problem-solving tasks under zero demonstration. 
For different methods, the same language model is used as base models.
To ensure a fair comparison, none of the methods used examples from the benchmark as demonstrations for in-context learning.

\paragraph{RAG Settings}
\method leverages the capabilities of Retrieval-Augmented Generation methods, which enhance the performance of language models by integrating external knowledge sources. Specifically, we employed the \texttt{codeparrot/github-jupyter} dataset as our primary search vector library for code generation and mathematical reasoning tasks. 
For embodied planning tasks in Minecraft, we utilized the Minecraft Wiki\footnote{\url{https://minecraft.wiki/}} and DigMinecraft\footnote{\url{https://www.digminecraft.com/}} websites as the information sources accessible to the LLMs.
For open-ended creative writing tasks, we use Google to search the query on the Internet.
We utilized OpenAI's \texttt{text-embedding-ada-002} API service for all embedding calculations across different methods and base models. 

Acknowledging the risk of benchmark contamination (an issue where the code library may contain solutions to the exact problems being evaluated), we adopted a rigorous pre-processing methodology as described by \citet{Guo2024DeepSeekCoderWT}. 
The potential implications of benchmark contamination, along with the effectiveness of our pre-processing strategy, are discussed in detail in Appendix~\ref{sec:benchmark_contamination}.

\subsection{Results}

The code generation results presented in Table~\ref{tab:code_generation} and results on other benchmarks presented in Table~\ref{tab:all} demonstrate the comprehensive evaluation of the \method across multiple benchmarks.
\method consistently outperforms the other methods across the majority of the benchmarks and metrics, showcasing its superior ability to generate long-horizon context. Notably, in the HumanEval and HumanEval+ benchmarks of code generation, \method achieves remarkable improvements in pass@1 and pass@5 rates, indicating a significant enhancement in first-attempt accuracy and within the top five attempts. For example, on the HumanEval benchmark, \method improves pass@1 by up to 20.94\% and pass@5 by up to 25.68\% relative to the base models' performances. This trend is observed across different underlying base models, highlighting \method's effectiveness regardless of the initial model's capabilities.
For mathematical reasoning tasks, \method demonstrates a significant relative improvement, with an 8.37\% increase in accuracy on GSM8K and a remarkable 31.37\% increase on GSMHard, culminating in an overall average improvement of 18.44\% when deployed on the GPT-3.5 model. 
\method significantly outperforms all other methods on open-ended embodied planning tasks in Minecraft, achieving the highest scores with 76.67±8.02\% for executability and 29.37 human rating score for plausibility, demonstrating its superior ability to generate feasible and contextually appropriate plans in the complex open-world environment. 
\method's superior performance also keeps across a broad spectrum of creative writing tasks. Its ability to generate high-quality content in diverse scenarios was demonstrated, highlighting its potential as a powerful tool for enhancing the general creative writing capabilities of LLMs in open-ended scenarios. 

The tasks are extremely diverse, while \method can have consistent improvements over all baselines. 
These results underline the advantages of \method's approach, which leverages iterative refinement of retrieval queries based on evolving reasoning thoughts. This strategy not only enhances the relevance and quality of the information retrieved but also significantly improves the accuracy and efficiency of the generated context.

\subsection{Case Analysis}

Here we take the embodied planning task and creative writing task to do case analysis.

In a manner analogous to multi-document question-answering tasks~\citep{Trivedi2022InterleavingRW}, the task of long-horizon planning in Minecraft is knowledge-dense, requiring consideration of various items for the completion of each task. 
However, open-world Minecraft knowledge on the internet is fragmented, making task completion often dependent on information from multiple sources. We observed that while language models like ChatGPT can identify necessary items through zero-shot CoT reasoning, inaccuracies in procedural steps are common. For example, ChatGPT inaccurately identified the materials for a crafting table as 4 wood blocks (the right answer is 4 planks), indicating lower executability reliability in CoT plans. Classical RAG algorithms, retrieving the knowledge with the question as a query and focusing on the final target item, inadequately retrieve intermediary items, offering minimal task improvement. Contrastingly, \method improves upon CoT's initial answers by continuously refining thoughts with targeted retrieval, aligning closely with task progression and relevant item knowledge. This methodology significantly enhances planning effectiveness by ensuring a comprehensive understanding and retrieval of all items involved in a plan, highlighting the synergy between structured reasoning and dynamic knowledge retrieval in addressing long-horizon planning challenges in Minecraft.

In addressing open-ended creative writing tasks, assessments of LM's generations typically focus on completeness and accuracy. When tasked with ``summarizing the American Civil War according to a timeline'', LMs under DIRECT and CoT prompts often produce significant hallucinations. For example, the statement ``The Civil War officially began on April 12, 1860, when Confederate troops attacked Fort Sumter in South Carolina, a Union-held fort'' contains incorrect information, where the year 1860 is erroneously mentioned instead of the correct year, 1861.

Direct queries to the internet for this task tend to retrieve limited events, frequently overlooking the accurate start date of the war, April 12, 1861. Moreover, the RAG approach, which tends to summarize content retrieved from searches, often misses this event in its responses, whether it's RAG-1 or RAG-5. On the other hand, \method bases its search on a language model's draft answer, finding that hallucinations usually occur in details, such as specific dates, which do not hinder the search engine from identifying relevant information like ``American Civil War starting date''. RAT utilizes the content retrieved to identify and correct errors in the draft answer rather than merely summarizing the retrieved content. Therefore, RAT can achieve a complete generation through reasoning and enhance the accuracy and credibility of the answer by leveraging retrieved knowledge. Experimental results validate the effectiveness of RAT.

\subsection{Ablation Study}\label{sec:exp_ablation}

\input{tables/ablation_retrieval}

\paragraph{Ablation on retrieval in \method}
In this ablation study, we investigate the influence of various retrieval strategies on the efficacy of \method, focusing on the optimization of content retrieval for improving generative outputs. The experimental results, detailed in Table~\ref{tab:ablation_on_retrieval}, highlight the significant advancements achieved through the iterative refinement of retrieval queries in \method compared to baseline methods. The baseline denoted as RAG-1, employs a direct approach by using the question itself as the retrieval query. In contrast, CoT+RAG enhances this process by utilizing the entirety of the reasoning thoughts output by the language model as the query, aiming for a broader contextual understanding. However, \method introduces a more dynamic method by employing continuously modified parts of reasoning thoughts as queries, which allows for a more focused and relevant information retrieval process.
The comparative analysis shows that \method surpasses both the baseline and the CoT+RAG method in terms of pass@1 and pass@5 metrics across the HumanEval and HumanEval+ benchmarks. Specifically, \method demonstrates an 8.7 percentage point increase in pass@1 and a 7.9 percentage point increase in pass@5 over the baseline in the HumanEval benchmark, and similarly impressive gains in the HumanEval+ benchmark. These improvements underscore the effectiveness of \method's retrieval strategy, which by iteratively refining next queries based on evolving reasoning thoughts and previous queries, ensures the retrieval of highly pertinent information. This process not only enhances the relevance of the information retrieved but also significantly improves the quality and accuracy of the final generated outputs. The results firmly establish the superiority of \method's dynamic retrieval method in leveraging contextual nuances to drive more precise and effective generative processes.


\input{tables/ablation_reason}

\paragraph{Ablation on causal reasoning in \method}
In this ablation study, we systematically examine the impact of causal and non-causal reasoning approaches on the performance of the \method system, with the Chain of Thought (CoT) serving as our baseline. Our findings, as summarized in Table~\ref{tab:ablation_on_reasoning}, reveal significant enhancements in generation capabilities when incorporating causal reasoning techniques. Specifically, the causal approach, which iteratively performs reasoning and retrieval, leads to notable improvements in both pass@1 and pass@5 metrics across HumanEval and HumanEval+ benchmarks. For instance, the causal method outperforms the baseline (CoT) by 11.9 percentage points in pass@1 and by 4.6 percentage points in pass@5 on the HumanEval dataset. This approach contrasts with the non-causal method, which, although also surpassing the baseline, leverages the initial reasoning thought to directly retrieve all necessary steps and generate the final answer. The causal method's superior performance underscores the value of sequential reasoning and information retrieval in enhancing the accuracy and reliability of generated outputs. This iterative process likely aids in refining the search and reasoning steps based on continuously updated context, allowing for more precise and relevant information retrieval, which in turn supports more accurate final answers. These results firmly establish the efficacy of causal reasoning in long-horizon problem-solving tasks.

\subsection{Robustness of \method}

\method was rigorously validated across a diverse set of tasks, including code generation, mathematical reasoning, creative writing, and embodied planning. This variety of tasks underscores the generalization capability of \method, demonstrating its robust performance across highly diverse challenges. Furthermore, all our experimental settings were conducted in a zero-shot manner; we did not design task-specific prompts for \method, but rather used the simplest possible prompts (which can be found in Appendix~\ref{sec:prompt}) to articulate questions or instructions for all methods. This approach ensures \method's generalization ability in open-ended scenarios.

The diversity of our evaluation was further enhanced by testing \method across various language models of differing capacities. This included CodeLlama-7b~\citep{codellama}, ChatGPT (gpt-3.5-turbo)~\citep{instructGPT}, and the more advanced GPT-4 (gpt-4) model~\citep{gpt4}. Remarkably, \method maintained its generalization capability across different scales of language models, showing improvements in benchmarks such as the HumanEval for code generation tasks. Notably, the largest improvement was observed with GPT-4, attributed to its superior ability for in-context learning from retrieved text. On MBPP+, CodeLlama-7b based \method has demonstrated performance degradation. This decline could be due to the limited in-context learning ability of smaller language models.

For mathematical reasoning tasks, \method demonstrated a significant relative improvement, with an overall average improvement of 18.44\% when applied to the GPT-3.5 model. This trend of improvement persisted with GPT-4, which achieved a remarkable 10.26\% relative improvement from DIRECT to \method. These findings highlight \method's robustness and its effective enhancement of language models' performance across a spectrum of computational and creative tasks.

\section{Related Works}

\paragraph{Retrieval-augmented Generation (RAG)}
Recently, RAG has gained popularity for boosting the performance of LLMs by guiding their generation process using the retrieved knowledge~\citep{Zhao2023RetrievingMI}. Without updating model parameters that may be expensive~\citep{lewis2020retrieval} or unstable~\citep{ke2022adapting,ke2022continual}, RAG is a cost-effective way for LLMs to interact with the external world~\citep{gu2018search,lewis2020retrieval}. RAG is widely applied to downstream tasks, such as code generation~\citep{zhou2022docprompting,lu2022reacc,nashid2023retrieval}, question answering~\citep{baek2023knowledge,siriwardhana2023improving}, and creative writing~\citep{wen2023grove,selfrag}. 

\paragraph{Reasoning-enhanced RAG} Some recent works also leverage reasoning to enhance the performance of RAG~\citep{chainofknowledge}. For example, IRCoT~\citep{trivedi2022interleaving} exploits CoT to generate better queries for retrieval, IRGR~\citep{ribeiro2022entailment}
performs iteratively retrieval to search for suitable
premises for multi-hop QA, GEEK~\citep{Liu2023GraduallyEE} can choose to query external knowledge or perform
a single logical reasoning step in long-horizon generation tasks, and ITRG~\citep{Feng2023RetrievalGenerationSA} performs retrieval based on the last-step generation. However, these previous RAG methods simply adopt a single query to retrieve the knowledge for question-answering tasks~\citep{gao2023retrieval,feng2023retrieval}, while our proposed \method performs retrieval using reasoning and draft answers in an autoregressive way, which significantly improves the performance of RAG in various tasks as demonstrated in Figure~\ref{fig:overall_results}.

\paragraph{Language Model for Reasoning}
The advancement of reasoning in language models has seen notable methodologies emerge since \ac{cot} was proposed by \citet{chainofthought}, which showcased LMs' ability to generate self-derived problem-solving strategies. This foundational work spurred further innovations such as the least-to-most prompting~\citep{zhou2022least}, zero-shot CoT~\citep{kojima2022large}, self-consistency~\citep{wang2022self}, zero-shot CoT without prompting~\citep{wang2024chain}. 
Moving beyond basic prompting, \citet{creswell2022selection} introduced the Selection-Inference framework, while \citet{zelikman2022star} developed STaR to refine reasoning through model finetuning. \citet{creswell2022faithful} proposed a faithful reasoning model, segmenting reasoning into dedicated steps, similar to Scratchpad's approach by \citet{nye2021show} for enhancing multi-step computation. Tree-of-Thought~\citep{treeofthought} and Graph-of-Thought~\citep{graphofthought} also expand the reasoning paths into a complex structure instead of linear CoT.
These methods usually aim to improve the reasoning ability of LLM by designing prompts or providing feedback from the environment to assist in better planning and decision-making~\citep{deps,react,reflexion,li2023chain,proagent}. 
However, \method takes a different approach by using RAG to access external knowledge that can help LLM with its reasoning process.

\section{Conclusion}
We have presented Retrieval Augmented Thoughts (RAT), a simple yet effective prompting strategy that synergies chain of thought (CoT) prompting and retrieval augmented generation (RAG) to address the challenging long-horizon reasoning and generation tasks. Our key ideas involve revising the zero-shot chain of thoughts produced by LLMs through RAG with the thoughts as queries, and causally revising the thoughts \& generating the response progressively. \method, a \textbf{zero-shot} prompting approach, has demonstrated significant advantages over vanilla CoT prompting, RAG, and other baselines on challenging code generation, mathematics reasoning, embodied task planning, and creative writing tasks. 

\section*{Acknowledgments}
We thank a grant from  CCF-Tencent Rhino-Bird Open Research Fund. One author is funded in part by NSF grants \#IIS-1943641, \#IIS-1956441, \#CCF-1837129, an SRA from Meta and a research gift from Amazon Alexa AI, and a gift from RelationalAI. 

%% file: tables/code_results.tex
\begin{table*}[t]
\centering
\caption{Code generation results on different benchmarks.$^*$All tests are evaluated under zero-shot (0-demonstration) settings. We also report the \textcolor{red}{relative improvements} between RAT and DIRECT methods.
}
\label{tab:code_generation}
\resizebox{0.99\linewidth}{!}{
\renewcommand\arraystretch{1.1}
\begin{tabular}{cccclcclcclcclcc}
\hline
                               &                          & \multicolumn{2}{c}{HumanEval}                                   & \multicolumn{1}{c}{} & \multicolumn{2}{c}{HumanEval+}                                  & \multicolumn{1}{c}{} & \multicolumn{2}{c}{MBPP}                                                          & \multicolumn{1}{c}{}    & \multicolumn{2}{c}{MBPP+}                                                         & \multicolumn{1}{c}{} & \multicolumn{2}{c}{Average~$\uparrow$}                                     \\ \cline{3-4} \cline{6-7} \cline{9-10} \cline{12-13} \cline{15-16} 
\multirow{-2}{*}{Base Models}  & \multirow{-2}{*}{Method} & pass@1                         & pass@5                         & \multicolumn{1}{c}{} & pass@1                         & pass@5                         & \multicolumn{1}{c}{} & pass@1                                  & pass@5                                  & \multicolumn{1}{c}{}    & pass@1                                  & pass@5                                  & \multicolumn{1}{c}{} & pass@1                         & pass@5                         \\ \hline
                               & DIRECT                   & {33.78\%}               & 40.85\%                        &                      & 30.85\%                        & 36.59\%                        &                      & 39.27\%                                 & 54.27\%                                 &                         & 41.22\%                                 & 48.17\%                                 &                      & 36.28\%                        & 44.97\%                        \\
                               & CoT                      & 27.86\%                        & 29.58\%                        &                      & 25.12\%                        & 27.83\%                        &                      & 31.99\%                                 & 55.91\%                                 &                         & \textbf{42.19\%}                        & 47.51\%                                 &                      & 31.79\%                        & 40.21\%                        \\
                               & RAG\_1 shot              & 37.50\%                        & 47.65\%                        &                      & 33.66\%                        & 41.83\%                        &                      & 35.41\%                                 & 51.63\%                                 &                         & 43.66\%                                 & 50.09\%                                 &                      & 37.56\%                        & 47.80\%                        \\
                               & RAG\_5 shot              & 38.90\%                        & 47.90\%                        &                      & 35.37\%                        & 42.75\%                        &                      & 34.06\%                                 & 53.90\%                                 &                         & 43.35\%                                 & \textbf{51.08\%}                        &                      & 37.92\%                        & 48.91\%                        \\
\multirow{-5}{*}{CodeLlama-7b} & \method                    & \textbf{39.57\%}               & \textbf{51.34\%}               &                      & \textbf{36.22\%}               & \textbf{46.50\%}               &                      & \textbf{40.86\%}                        & \textbf{60.63\%}                        &                         & 39.14\%                                 & 48.04\%                                 &                      & \textbf{38.95\%}               & \textbf{51.63\%}               \\ \hline
\multicolumn{2}{c}{Relative Improvement}                  & {\color[HTML]{FF0000} 17.14\%} & {\color[HTML]{FF0000} 25.68\%} &                      & {\color[HTML]{FF0000} 17.41\%} & {\color[HTML]{FF0000} 27.08\%} &                      & {\color[HTML]{FF0000} 4.05\%}           & {\color[HTML]{FF0000} 11.72\%}          &                         & {\color[HTML]{FF0000} -5.05\%}          & {\color[HTML]{FF0000} -0.27\%}          &                      & {\color[HTML]{FF0000} 7.35\%}  & {\color[HTML]{FF0000} 14.80\%} \\ \hline
                               & DIRECT                   & {50.49\%}               & 72.56\%                        &                      & 48.09\%                        & 70.55\%                        &                      & \textbf{60.84\%}                        & 72.95\%                                 &                         & 54.92\%                                 & 64.09\%                                 &                      & 53.59\%                        & 70.04\%                        \\
                               & CoT                      & 47.31\%                        & 75.88\%                        &                      & 41.72\%                        & 74.85\%                        &                      & 55.19\%                                 & 65.49\%                                 &                         & 47.69\%                                 & 62.94\%                                 &                      & 47.98\%                        & 69.79\%                        \\
                               & RAG\_1 shot              & 50.61\%                        & 76.22\%                        &                      & 48.22\%                        & 70.55\%                        &                      & 55.23\%                                 & 70.54\%                                 &                         & 53.62\%                                 & 68.09\%                                 &                      & 51.92\%                        & 71.35\%                        \\
                               & RAG\_5 shot              & 45.49\%                        & 74.39\%                        &                      & 42.58\%                        & 70.55\%                        &                      & 54.39\%                                 & 69.73\%                                 &                         & 55.98\%                                 & 70.10\%                                 &                      & 49.61\%                        & 71.19\%                        \\
\multirow{-5}{*}{GPT-3.5}      & \method                    & \textbf{59.27\%}               & \textbf{80.49\%}               &                      & \textbf{56.31\%}               & \textbf{76.07\%}               &                      & {\color[HTML]{000000} {59.31\%}} & {\color[HTML]{000000} \textbf{74.74\%}} & {\color[HTML]{000000} } & {\color[HTML]{000000} \textbf{59.10\%}} & {\color[HTML]{000000} \textbf{72.61\%}} &                      & \textbf{58.50\%}               & \textbf{75.98\%}               \\ \hline
\multicolumn{2}{c}{Relative Improvement}                  & {\color[HTML]{FF0000} 17.39\%} & {\color[HTML]{FF0000} 10.93\%} &                      & {\color[HTML]{FF0000} 17.09\%} & {\color[HTML]{FF0000} 7.82\%}  &                      & {\color[HTML]{FF0000} -2.51\%}          & {\color[HTML]{FF0000} 2.45\%}           &                         & {\color[HTML]{FF0000} 7.61\%}           & {\color[HTML]{FF0000} 13.29\%}          &                      & {\color[HTML]{FF0000} 9.17\%}  & {\color[HTML]{FF0000} 8.48\%}  \\ \hline
                               & DIRECT                   & 57.32\%                        & 78.66\%                        &                      & 54.36\%                        & 76.69\%                        &                      & 60.00\%                                 & 76.07\%                                 &                         & 66.13\%                                 & 78.53\%                                 &                      & 59.45\%                        & 77.49\%                        \\
                               & CoT                      & 54.87\%                        & 72.56\%                        &                      & 51.90\%                        & 66.25\%                        &                      & 61.22\%                                 & 74.23\%                                 &                         & 64.42\%                                 & 79.75\%                                 &                      & 58.10\%                        & 73.20\%                        \\
                               & RAG\_1 shot              & 61.10\%                        & 79.27\%                        &                      & 58.04\%                        & 77.30\%                        &                      & 58.53\%                                 & 69.94\%                                 &                         & 65.77\%                                 & 77.30\%                                 &                      & 60.86\%                        & 75.95\%                        \\
                               & RAG\_5 shot              & 62.80\%                        & 82.93\%                        &                      & 59.51\%                        & 79.75\%                        &                      & 60.12\%                                 & 74.23\%                                 &                         & 63.56\%                                 & 78.53\%                                 &                      & 61.50\%                        & 78.86\%                        \\
\multirow{-5}{*}{GPT-4}        & \method                    & \textbf{69.33\%}               & \textbf{88.40\%}               &                      & \textbf{64.63\%}               & \textbf{82.21\%}               &                      & \textbf{68.90\%}                        & \textbf{79.85\%}                        &                         & \textbf{67.36\%}                        & \textbf{82.14\%}                        &                      & \textbf{67.55\%}               & \textbf{83.15\%}               \\ \hline
\multicolumn{2}{c}{Relative Improvement}                  & {\color[HTML]{FF0000} 20.94\%} & {\color[HTML]{FF0000} 12.38\%} &                      & {\color[HTML]{FF0000} 18.89\%} & {\color[HTML]{FF0000} 7.20\%}  &                      & {\color[HTML]{FF0000} 14.83\%}          & {\color[HTML]{FF0000} 4.97\%}           &                         & {\color[HTML]{FF0000} 1.86\%}           & {\color[HTML]{FF0000} 4.60\%}           &                      & {\color[HTML]{FF0000} 13.63\%} & {\color[HTML]{FF0000} 7.31\%}  \\ \hline
\end{tabular}}
\vskip -0.1in
\end{table*}

%% file: tables/merge_table.tex

\begin{table*}[]
\centering
\caption{\textbf{Evaluation results on mathematical reasoning, creative writing, and embodied planning tasks.} Among them, mathematical reasoning and creative writing use \texttt{gpt-3.5} as base models, while embodied planning uses \texttt{gpt-4} as base models. $\Delta$ represents the relative improvements than DIRECT.}
\label{tab:all}
\resizebox{0.99\linewidth}{!}{
\renewcommand\arraystretch{1.1}
\begin{tabular}{@{}lccccccccccc@{}}
\toprule
\multirow{2}{*}{Method} & \multicolumn{3}{c}{\textbf{Math Reasoning Accuracy~$^*$}~$\uparrow$}                    &  & \multicolumn{3}{c}{\textbf{Creative Writing}~$\uparrow$}                        &  & \multicolumn{3}{c}{\textbf{Embodied Planning}~$\uparrow$}                          \\ \cmidrule(lr){2-4} \cmidrule(lr){6-8} \cmidrule(l){10-12} 
                        & GSM8K            & GSMHard          & Average~($\Delta$)        &  & Win Rate         & TrueSkill Rating~($\Delta$) & Uncertainty &  & Executablity          & Plausibitlity~($\Delta$)  & Uncertainty \\ \midrule
DIRECT                  & 65.85\%          & 51.26\%          & 58.56\%                  &  & 46.67\%          & 24.39                      & 1.17        &  & 19.33±2.08\%          & 20.57                    & 2.05        \\
CoT                     & 63.82\%          & 44.72\%          & 54.27(-7.32)\%           &  & 41.67\%          & 24.31(-0.0\%)              & 1.09        &  & 49.33±3.05\%          & 25.75(+25.2\%)           & 2.33        \\
RAG-1 shot              & 61.81\%          & 51.26\%          & 56.54(+4.17)\%           &  & 38.71\%          & 23.99(-1.6\%)              & 1.11        &  & 31.00±5.29\%          & 24.97(+21.4\%)           & 2.11        \\
RAG-5 shot              & 61.81\% & 56.78\% & 59.30(+4.88)\%           &  & 31.67\%          & 23.88(-2.1\%)              & 1.22        &  & 33.00±3.61\%          & 25.02(+21.6\%)           & 2.11        \\
RAT                   & \textbf{71.36\%} & \textbf{67.34\%} & \textbf{69.35(+16.96)\%} &  & \textbf{81.01\%} & \textbf{29.07(+19.2\%)}     & 1.08        &  & \textbf{76.67±8.02\%} & \textbf{29.37(+42.78\%)} & 3.37        \\ \bottomrule
\end{tabular}}
\vskip -0.2in
\end{table*}

%% file: tables/ablation_retrieval.tex
\begin{table}[]
\centering
\caption{Comparative Impact of Retrieval Strategies on \method Performance.}
\label{tab:ablation_on_retrieval}
\resizebox{0.99\linewidth}{!}{
\renewcommand\arraystretch{1.2}
\begin{tabular}{@{}lcclcc@{}}
\toprule
\multirow{2}{*}{Method} & \multicolumn{2}{c}{HumanEval} &  & \multicolumn{2}{c}{HumanEval+}  \\ \cmidrule(lr){2-3} \cmidrule(l){5-6} 
                        & pass@1($\Delta$)$~\uparrow$ & pass@5($\Delta$)$~\uparrow$ &  & pass@1($\Delta$)$~\uparrow$  & pass@5($\Delta$)$~\uparrow$  \\ \midrule
Baseline         & 50.6\%       & 76.2\%       &  & 48.2\%        & 70.5\%        \\
CoT+RAG                 & 53.9({\color{red}+3.3})\% & 76.8({\color{red}+0.6})\% &  & 51.3({\color{red}+3.1})\% & 69.3({\color{red}-1.2})\% \\
\method                   & 59.2({\color{red}+8.7})\% & 80.4({\color{red}+7.9})\% &  & 56.3({\color{red}+8.2})\%  & 76.0({\color{red}+5.5})\%  \\ \bottomrule
\end{tabular}}
\vskip -0.2in
\end{table}

%% file: tables/ablation_reason.tex
\begin{table}[]
\centering
\caption{Ablation Study on Causal vs. Non-Causal Reasoning in \method.}
\label{tab:ablation_on_reasoning}
\resizebox{0.99\linewidth}{!}{
\renewcommand\arraystretch{1.2}
\begin{tabular}{@{}lcclcc@{}}
\toprule
\multirow{2}{*}{Method} & \multicolumn{2}{c}{HumanEval} &  & \multicolumn{2}{c}{HumanEval+}  \\ \cmidrule(lr){2-3} \cmidrule(l){5-6} 
                        & pass@1($\Delta$)$~\uparrow$ & pass@5($\Delta$)$~\uparrow$ &  & pass@1($\Delta$)$~\uparrow$  & pass@5($\Delta$)$~\uparrow$  \\ \midrule
Baseline           & 47.3\%        & 75.8\%        &  & 41.7\%        & 74.8\%       \\
Non-Causal              & 57.3({\color{red}+10.0})\% & 78.0({\color{red}+2.1})\% &  & 54.9({\color{red}+13.2})\% & 74.8({\color{red}+0.0})\%   \\
Causal                  & 59.2({\color{red}+11.9})\% & 80.4({\color{red}+4.6})\%  &  & 56.3({\color{red}+14.6})\% & 76.0({\color{red}+1.2})\% \\ \bottomrule
\end{tabular}}
\vskip -0.2in
\end{table}

%% file: appendix.tex
\onecolumn

\section{Task Details}

\subsection{Code Generation}

\paragraph{Benchmarks} 
We select HumanEval~\citep{codex}, HumanEval+~\citep{evalplus}, MBPP~\citep{mbpp}, and MBPP+~\citep{evalplus} as the code generation evaluation benchmark. 
These benchmarks are commonly used to test the performance of code generation models, which are briefly introduced below:
\begin{itemize}
 \item \textbf{HumanEval} consists of 164 Python programming problems, each with a function signature, docstring, body, and multiple unit tests~\citep{codex}.
 \item \textbf{HumanEval+} includes the same programming problems as HumanEval, but with an additional 80 times more unit tests for each of the 164 problems~\citep{evalplus}. 
 \item \textbf{MBPP} is a collection of approximately 1,000 Python programming problems that are intended to be solvable by beginner programmers. Each problem includes an English task description, a code solution, and three automated test cases. We assess the sample test set from index 11 to 175~\citep{mbpp}.
 \item \textbf{MBPP+} consists of 399 tasks~\citep{evalplus}, which are a subset of the original MBPP dataset. Additionally, MBPP+ includes extra unit tests for each of the 399 problems (35 times more than the original MBPP). We utilized the first 164 questions as our test set.
\end{itemize}
These benchmarks encompass a wide range of programming problems, from simple function implementations to more complex algorithmic challenges, providing a robust testbed for assessing the generative capabilities of various models.

\paragraph{Metrics}
We adopt the \texttt{pass@k} metric for evaluating the efficacy of various code generation algorithms, following the methodology proposed by \citet{codex} and extended by \citet{evalplus}. This metric quantifies the rate at which generated code snippets successfully execute and pass all test cases, where $k$ represents the number of attempts or samples generated by the model for each problem. 
This approach allows us to rigorously assess the precision and reliability of code generation models in producing functionally correct code across a diverse set of programming challenges.

\paragraph{Baselines}
To establish a comprehensive and equitable comparison landscape, we incorporate a suite of baseline methods and diverse code generation models. Our baselines include the original code generation language models, referred to as DIRECT, and the Retrieval-Augmented Generation (RAG) methodology with $n$ retrieved examples, instantiated in both single-shot (1 shot) and multi-shot (5 shots) configurations, as documented by \citet{rag}. Additionally, we examine the zero-shot \ac{cot} (CoT) approach, as conceptualized by \citet{kojima2022large}, which simulates a step-by-step reasoning process to facilitate complex problem-solving tasks under zero demonstration. 
To ensure a fair comparison, none of the methods used examples from the benchmark as demonstrations for in-context learning.

The diversity of our evaluation is further enriched by testing across various language models with differing capacities, including CodeLlama-7b~\citep{codellama}, along with ChatGPT(\texttt{gpt-3.5-turbo})~\citep{instructGPT}, and the more advanced GPT-4(\texttt{gpt-4}) model~\citep{gpt4}.
Recognizing the potential format discrepancies in code outputs, especially considering that models like \texttt{gpt-3.5-turbo} and \texttt{gpt-4} may produce code in markdown format which is not immediately executable, we implement post-processing steps to convert the original language model outputs into a form that can be executed within a sandbox environment. This normalization ensures that all models are evaluated under uniform execution conditions, thereby facilitating a fair and direct comparison of their code generation capabilities. Through this methodological framework, we aim to provide a detailed and nuanced understanding of the performance landscape across a spectrum of LLM-driven code generation approaches.

\paragraph{RAG Settings} 
\method leverages the capabilities of Retrieval-Augmented Generation methods, which enhance the performance of language models by integrating external knowledge sources.
Specifically, we employed the \texttt{codeparrot/github-jupyter} dataset as our primary search vector library. This dataset is a comprehensive compilation of 452k markdown and code pairs, meticulously extracted from Jupyter notebooks hosted on GitHub BigQuery, representing a rich repository of programming knowledge and examples.
We utilized OpenAI's \texttt{text-embedding-ada-002} API service for all embedding calculations across different methods and base models. 

\subsection{Mathematical Reasoning}

\paragraph{Benchmarks} Our evaluation framework for assessing mathematical reasoning capabilities leverages two primary benchmarks: the GSM8K dataset, which comprises over 8,000 multi-step mathematical problems~\citep{gsm8k}, and the GSM-HARD dataset, an adaptation of GSM8K where numbers in the questions are replaced with larger values to increase problem complexity~\citep{pal}. This study employs the PAL methodology to scrutinize the mathematical reasoning results, involving the utilization of Large Language Models (LLMs) to parse natural language problems, generate intermediary programmatic solutions, and subsequently execute these solutions via a Python interpreter. The test set for each benchmark consists of samples ranging from index 1 to 200. Uniquely, our approach does not use any examples for in-context learning, differing from the original PAL methods. 

\paragraph{Metrics and Baselines} 
Accuracy serves as our principal metric for evaluation, aligning with the established metric for the GSM8K benchmark. Each question undergoes three execution attempts, with the average score recorded as the final result. The baselines, including DIRECT, CoT, RAG (1 shot), and RAG (5 shots), are consistent with those outlined in code generation, facilitating a comprehensive and comparative analysis across different code generation benchmarks. The RAG settings are consistent with the code generation tasks.

\subsection{Embodied Planning}

We further conduct experiments on embodied planning benchmarks on open-ended environments Minecraft~\citep{mcu}.

\paragraph{Benchmarks}
The complexity and vast item interconnectivity within the open-world Minecraft present an ideal testbed for evaluating the LLM's capability to generate long-horizon plans~\citep{plan4mc,deps,jarvis1}. With thousands of items and intricate relationships between them, obtaining a specific item in survival mode from scratch may involve dozens of intermediate items and their quantitative relationships, such as crafting 1 crafting table from 4 planks. This setting rigorously tests the planning abilities of LLMs instead of low-level control policies~\citep{groot,vpt,Cai2023OpenWorldMC,steve1,yuan2024pretraining}. Moreover, \citet{jarvis1} have identified instances of hallucinations about Minecraft knowledge in OpenAI's ChatGPT and a general scarcity of Minecraft-related knowledge in open-source language models, making this task a suitable benchmark for assessing the RAG algorithm's effectiveness.

The planning prompts are aligned with those used in DEPS~\citep{deps}, structured as Python templates and evaluated using MC-TextWorld as detailed by \citet{mcu}. 
A set of 100 tasks were randomly selected for the test set, ranging from simple objectives like obtaining a crafting table to more complex goals such as crafting an iron helmet and even challenging making an enchanting table.
The task instruction is formulated as:
\begin{itemize}
    \item Give you nothing in the inventory, generate a step-by-step plan for the task of obtaining a \{placeholder:acacia\_boat\} in Minecraft survival mode, and describe the object Minecraft item and its number at every step. For every step, start with 'STEP' as start.
    \item Give you nothing in the inventory, generate a step-by-step plan for the task of obtaining a \{placeholder:diamond\_pickaxe\} boat in Minecraft survival mode, and describe the object Minecraft item and its number at every step. For every step, start with 'STEP' as start.
\end{itemize}
There are over 100 tasks involving different Minecraft items.

\paragraph{RAG Settings}
For the retrieval component of the RAG algorithm, we utilized the Minecraft Wiki\footnote{\url{https://minecraft.wiki/}} and DigMinecraft\footnote{\url{https://www.digminecraft.com/}} websites as the information sources accessible to the LLMs. Data from these websites was cleaned and formatted into markdown text, then segmented into trunks not exceeding 2000 tokens each, with embedding calculations performed using OpenAI's \texttt{text-embedding-ada-002} API service.

\paragraph{Evaluation Metrics}
Based on the methodology of \citet{huang2022language}, our evaluation of open-ended, long-horizon planning in Minecraft focuses on both executability and plausibility.
Executability primarily examines whether a plan can be carried out, including the accuracy of each step's preconditions and effects. The executability is automatically calculated using MC-TextWorld~\citep{mcu}. However, executability only evaluates if an objective-level plan can be executed, without considering the specific details involved in executing individual objectives. For instance, crafting a wooden pickaxe requires placing a crafting table and arranging three planks and two sticks in a particular pattern, which are important details for human execution but not assessed by MC-TextWorld. Therefore, we complement our evaluation with human ratings to assess the plausibility of plans.

\subsection{Creative Writing}

To further understand the potential of Retrieval-Augmented Generation (RAG) models in enhancing the creativity and relevance of generated content, we extend our investigation to open-ended text generation tasks within the realm of creative writing.

\paragraph{Benchmarks}
The versatility of \method was tested through a series of creative writing tasks, each chosen to highlight different aspects of open-ended text generation. These tasks include:
\begin{itemize}
\item Write a survey paper to summarize the \texttt{placeholder:Retrieval-augmented Generation methods for Large Language Models}. 
\item Describe of \texttt{placeholder:Jin-Yong}'s life.
\item Summarize the \texttt{placeholder:American Civil War} according to the timeline.
\end{itemize}
For each task, three variants for \texttt{placeholder} were created to ensure a comprehensive evaluation of the model's performance across different contexts and requirements. 

\paragraph{RAG Settings}
Differing from previous tasks, creative writing is categorized as an open-ended generation task, demanding a broader scope of information retrieval to aid content generation. To accommodate this, Google was utilized as the search engine, with the top-k web pages converted into markdown text to assist the LLM in generating outputs. 
This approach allowed LLM to leverage a wide array of information sources.

\paragraph{Baselines and Evaluations}
To benchmark \method's performance, we compared it against DIRECT, RAG-1 shot, and RAG-5 shot methods, all based on the \texttt{gpt-3.5-turbo} model. The evaluation was conducted by human experts, employing the TrueSkill rating system~\citep{trueskill} to calculate scores for each method. This evaluation framework enabled a comprehensive assessment of each model's creative output quality, accuracy, relevance, and innovativeness.

\section{Prompt Details}\label{sec:prompt}

Our prompts consist of three parts: prompt for generating initial answer, prompt for generating search query, and prompt for revising answers according to retrieved context. 

\begin{prompt}[title={Prompt \thetcbcounter: Prompt for generating initial answers in creative writing tasks}, label=creative_draft_prompt]

\{\textcolor{blue}{\textbf{user}}\}\\
\#\#Question: \\
\textbf{\{question\}}\\
\#\#Instruction: \\
Try to answer this question/instruction with step-by-step thoughts and make the answer more structural.\\
Use /n/n to split the answer into several paragraphs.\\
Just respond to the instruction directly. DO NOT add additional explanations or introducement in the answer unless you are asked to.

\{\textcolor{blue}{\textbf{assistant}}\}\\
...
\end{prompt}

The process of query generation is omitted in code generation tasks. Instead, we use the generated code draft as a query and compute the embedding of it based on OpenAI Embedding services. For embodied planning and creative writing tasks, we will generate an additional query. 

\begin{prompt}[title={Prompt \thetcbcounter: Prompt for generating open-search query in creative writing tasks}, label=creative_query_prompt]

\#\#Question: \\
\textbf{\{question\}} \\
\#\#Content: \\
\textbf{\{answer\}} \\
\#\#Instruction: \\
I want to verify the content correctness of the given question, especially the last sentences.\\
Please summarize the content with the corresponding question.\\
This summarization will be used as a query to search with Bing search engine.\\
The query should be short but need to be specific to promise Bing can find related knowledge or pages.\\
You can also use search syntax to make the query short and clear enough for the search engine to find relevant language data.\\
Try to make the query as relevant as possible to the last few sentences in the content.\\
**IMPORTANT**\\
Just output the query directly. DO NOT add additional explanations or introducement in the answer unless you are asked to.\\
\{\textcolor{blue}{\textbf{assistant}}\}\\
...

\end{prompt}

\begin{prompt}[title={Prompt \thetcbcounter: Prompt for revising answer according to retrieved materials in creative writing tasks}, label=creative_revise_prompt]

\{\textcolor{blue}{\textbf{user}}\}\\
\#\#Existing Text in Wiki Web: \\
\textbf{\{content\}} \\
\#\#Question: \\
\textbf{\{question\}} \\
\#\#Answer: \\
\textbf{\{answer\}} \\
\#\#Instruction: \\
I want to revise the answer according to retrieved related text of the question in WIKI pages.\\
You need to check whether the answer is correct.\\
If you find some errors in the answer, revise the answer to make it better.\\
If you find some necessary details are ignored, add it to make the answer more plausible according to the related text. \\
If you find the answer is right and do not need to add more details, just output the original answer directly. \\
**IMPORTANT** \\
Try to keep the structure (multiple paragraphs with its subtitles) in the revised answer and make it more structural for understanding. 
Split the paragraphs with /n/n characters.
Just output the revised answer directly. DO NOT add additional explanations or annoucement in the revised answer unless you are asked to.

\{\textcolor{blue}{\textbf{assistant}}\}\\
...

\end{prompt}

\section{TrueSkill Evaluation Framework}

Part of the tasks in ``Embodied planning'' and ``creative writing'' involve using humans for labeling. Human labelers have 4 choices: ``A is better'', ``B is better'', ``Tie'' or ``Both are bad''. In this case, ``Tie'' and ``Both are bad'' will be counted as a tie.
For each task group, we have selected more than 10 professional annotators to provide labels.
We use the Python ``trueskill'' package to calculate the win rate and score. The default score for every method is set as 25.
In order to facilitate user understanding and selection, we also provide prompts when entering the system.

\begin{prompt}[colback=red!5!white,colframe=blue!75!black]
\#  Chatbot Arena : Benchmarking LLMs in the Wild\\
\#\#Rules\\
- Refresh to obtain the question and its corresponding answers from two anonymous models.\\
- Vote for the better answer. And then click "New Round" to get a new question.\\
- If both answers are bad, vote for "Both are bad".\\
- If you want to skip, click "Skip".\\

\#\# Principle\\
You can evaluate the performance of the model from the following aspects:\\
1. **Relevance**: Does it answer the question accurately?\\
2. **Accuracy**: Is it accurate? For example, a crafting table is made by combining 4 wooden planks, not 4 logs; a diamond axe requires 3 diamonds and 2 sticks to craft, not 3 sticks and 2 diamonds.\\
3. **Completeness**: Is it complete? For example, crafting a wooden pickaxe from logs requires first crafting wooden planks and then crafting sticks before finally being able to craft the pickaxe. The intermediate steps cannot be ignored.\\
4. **Readability**: Is it coherent?\\
5. **Executability**: Considering the characteristics of the game, is it executable?\\

\#\# Vote now!\\
\end{prompt}

\begin{table*}[t]
    \centering
    \includegraphics[width=0.99\linewidth]{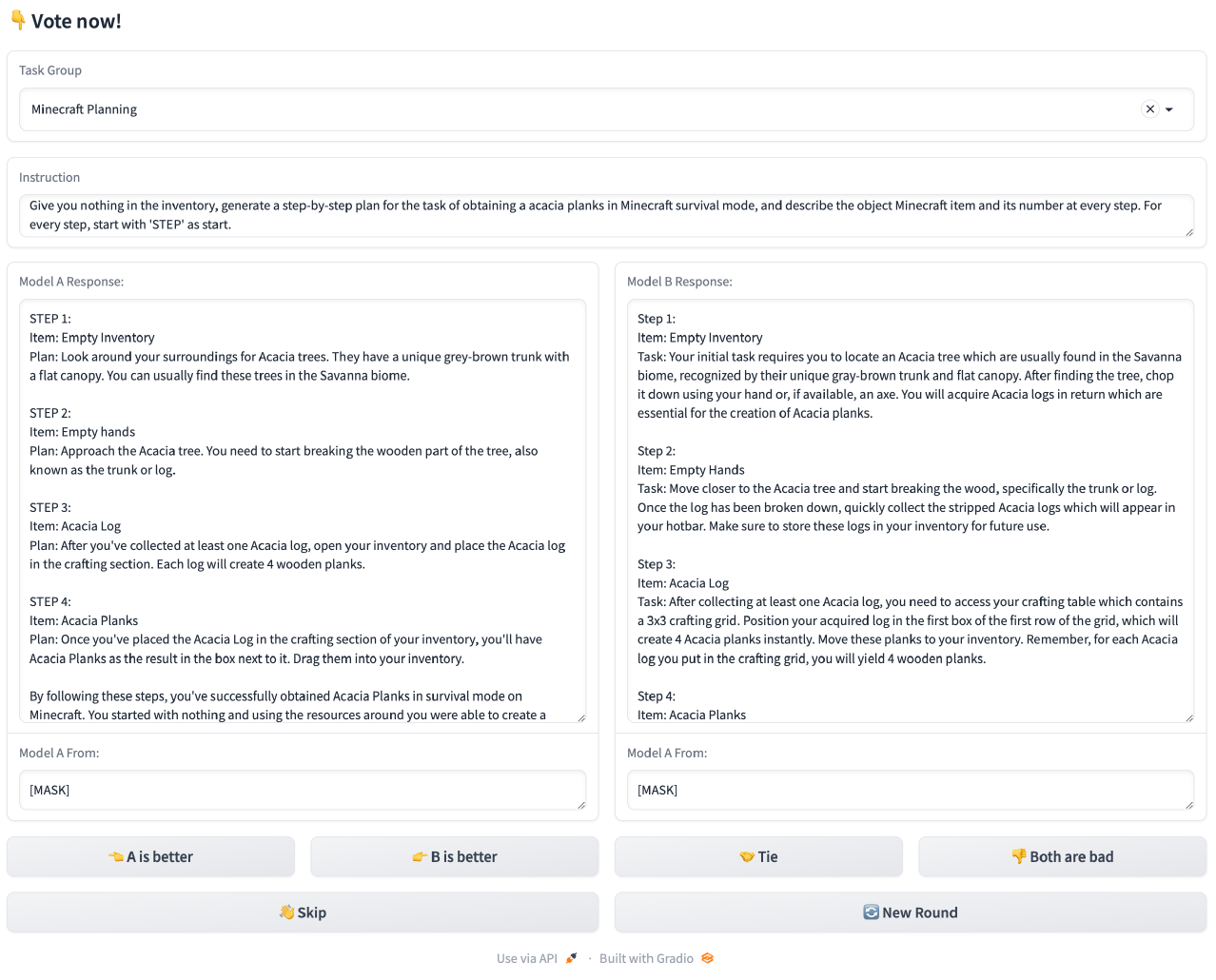}
    \captionof{figure}{
        The human evaluation UI. 
        We will display responses from two different methods for the same instruction on the page simultaneously. The source of the response will be marked as [MASK], and after human labeling, [MASK] will be replaced with the specific method name.
    }
    \label{fig:human_eva_ui}
\end{table*}

\section{Disscussions on Benchmark Contamination}\label{sec:benchmark_contamination}
To avoid the code library containing solutions to the exact problems being evaluated) in code generation benchmarks, we adopted a rigorous pre-processing methodology as described by \citet{Guo2024DeepSeekCoderWT}. 
This process was meticulously designed to remove any direct matches or overly similar code snippets from our search vector library, thereby ensuring that our evaluation remains fair and uncontaminated by pre-existing solutions.
This examination aims to underscore the importance of maintaining the integrity of the evaluation process while utilizing external knowledge sources to augment the capabilities of language models in code-generation tasks.

\input{tables/contamination}

To further explore the potential benchmark contamination, we also conducted additional finetuning on CodeLLaMA-7B-Python using the code corpus in \autoref{tab:contamination}.

\section{More Results}

\subsection{Emboddied Planning}

Embodied planning involves multiple steps, each of which relies on specific world knowledge and causal knowledge (i.e., preceding steps are usually preconditions for subsequent steps), including recipes for items in Minecraft, tools for performing related actions, and quantity restrictions. Therefore, although the plan generated by ChatGPT may appear complete and correct, there are often errors within the steps that can affect the proper execution of the plan.

We mark the errors found in the generated plan in \textcolor{red}{red}.

\begin{prompt}[title={Response \thetcbcounter: Response from Zero-Shot CoT(based on ChatGPT-3.5) in embodied planning tasks}, label=embodied_original_response]

\{\textcolor{blue}{\textbf{instruction}}\}\\
Give you nothing in the inventory, generate a step-by-step plan for the task of obtaining a golden apple in Minecraft survival mode, and describe the object Minecraft item and its number at every step. For every step, start with 'STEP' as start.\\

\{\textcolor{blue}{\textbf{response}} from Original ChatGPT\}\\
STEP 1: Start by punching trees to gather at least 4 logs. 
- Minecraft item: 4x Logs \\
STEP 2: \textcolor{red}{Craft the logs into a crafting table.} 
- Minecraft item: 1x Crafting Table\\
STEP 3: Place the crafting table on the ground and right-click on it to open the crafting interface.\\
STEP 4: Use the crafting interface to \textcolor{red}{turn 4 of the oak planks into a wooden pickaxe}.\\ 
- Minecraft item: Wooden Pickaxe\\
STEP 5: Find a cave or dig into the ground to gather some stone. Mine at least 8 blocks of stone using your wooden pickaxe. 
- Minecraft item: 8x Cobblestone\\
STEP 6: Craft the cobblestone into a furnace using the crafting table. 
- Minecraft item: 1x Furnace \\
STEP 7: Place the furnace on the ground and right-click on it to open the furnace interface.\\
STEP 8: \textcolor{red}{Use the furnace interface to smelt some iron ore.} Place at least 3 iron ore in the top slot of the furnace and use wooden planks as fuel in the bottom slot. 
- Minecraft item: 3x Iron Ingots\\
STEP 9: Craft the iron ingots into a bucket using the crafting table. 
- Minecraft item: 1x Bucket\\
STEP 10: Find a water source (such as a lake or a river) and right-click on it while holding the bucket to fill it with water. 
- Minecraft item: 1x Water Bucket\\
STEP 11: Look for an apple tree in a forest biome. Break some leaves until you find an apple. 
- Minecraft item: 1x Apple\\
STEP 12: \textcolor{red}{Combine the water bucket and the apple in the crafting table to craft a golden apple}. 
- Minecraft item: 1x Golden Apple\\
Congratulations! You have successfully obtained a golden apple in Minecraft survival mode.

\end{prompt}

Although the Zero-shot CoT has generated a step-by-step plan overall, there are many factual errors within it. These include recipe errors in STEP 2, where the crafting table requires planks instead of wood; missing raw materials in STEP 4, as the wooden pickaxe needs 2 sticks but lacks the relevant step in the plan; absence of instructions to use a stone pickaxe to mine iron ore in STEP 8; and an incorrect recipe for golden apple in STEP 12, which should include gold ingots and an apple rather than a water bucket.

\begin{prompt}[title={Response \thetcbcounter: Response from RAT(based on ChatGPT-3.5) in embodied planning tasks}, label=embodied_rat_response]

\{\textcolor{blue}{\textbf{instruction}}\}\\
Give you nothing in the inventory, generate a step-by-step plan for the task of obtaining a golden apple in Minecraft survival mode, and describe the object Minecraft item and its number at every step. For every step, start with 'STEP' as start.\\

\{\textcolor{blue}{\textbf{response}} from Original ChatGPT\}\\

STEP 1: Find a tree and chop it down to get wood. You need at least 4 Oak Logs.\\
- Minecraft items: 4x Oak Logs\\
STEP 2: Craft the oak log into oak planks.\\
- Minecraft items: 16x Oak Planks\\
STEP 3: Convert some oak planks into sticks.\\
- Minecraft items: 4x Stick\\
STEP 4: Use the crafting table to create a wooden pickaxe.\\
- Minecraft items: 1x Wooden Pickaxe\\
STEP 5: Start digging down to find stone and gather at least three pieces of cobblestone.\\
- Minecraft items: 3x Cobblestone\\
STEP 6: Use the crafting table to create a stone pickaxe.\\
- Minecraft items: 1x Stone Pickaxe\\
STEP 7: Explore caves and mineshafts to find iron ore. Collect 3 iron ore for crafting an iron pickaxe.\\
- Minecraft items: 3x Iron Ore\\
STEP 8: Smelt the iron ore \textcolor{red}{using a furnace} to obtain iron ingot.\\
- Minecraft items: 3x Iron Ingot\\
STEP 9: Use the crafting table to create an iron pickaxe.\\
- Minecraft items: 1x iron pickaxe\\
STEP 10: Explore caves and mineshafts to find gold ore. Use an iron pickaxe or higher to mine the gold ore block. Collect as much gold ore as needed for crafting a golden apple.\\
- Minecraft items: 8x Gold Ore\\
STEP 11: Smelt the raw gold using a furnace to obtain a gold ingot.\\
- Minecraft items: 8x Gold Ingot\\
STEP 12: Find an oak tree and chop it down until an apple falls.\\
- Minecraft items: 1x Apple\\
STEP 13: Open the crafting menu and place the items in the correct pattern to craft a Golden Apple. In the 3x3 crafting grid, place 8 Gold Ingots in the first and third rows, leaving the middle square empty. In the second row, place 1 Gold Ingot in the first and third squares, and place an Apple in the middle square.\\
- Minecraft items: 1x Golden Apple\\
Congratulations! You have successfully obtained a golden apple in Minecraft survival mode.

\end{prompt}

There are still errors in the plan generated by RAT, such as the lack of tools before "STEP 8: smelt iron ore into iron ingots", which should be ``Mine 8 cobblestone to craft 1 furnace''. However, compared to the errors in ChatGPT, the error rate in the plan has been significantly reduced. 

We have also listed the links of the retrieved pages involved in different steps. 
We can see that the text sources retrieved in each step generated by RAT are usually highly related to the synthesized item of that step. Traditional RAG uses instructions for retrieval and can only find the final step. 
While RAT can retrieve all links related to intermediate items, which improves the accuracy and plausibility greatly. 

\input{tables/minecraft_links}

\subsection{Creative Writing}

\begin{prompt}[title={Response \thetcbcounter: Response from ORIGINAL ChatGPT-3.5 in creative writing tasks}, label=creative_original_response]

\{\textcolor{blue}{\textbf{instruction}}\}\\
Describe the history of the United States.
\{\textcolor{blue}{\textbf{response}} from Original ChatGPT\}\\

The history of the United States dates back thousands of years, with indigenous peoples such as the Native Americans inhabiting the land long before the arrival of European settlers. However, the official history of the United States begins in 1776 when the Founding Fathers declared independence from Great Britain. This marked the start of the American Revolution, a conflict that ultimately led to the creation of the United States as an independent nation.\\
After gaining independence, the United States went through a period of rapid expansion in the 19th century. The country acquired vast amounts of territory, including the Louisiana Purchase and the annexation of Texas. This expansion was accompanied by conflicts, such as the Mexican-American War and the Civil War, which tested the unity of the nation.\\
In the 20th century, the United States emerged as a global superpower. The country played a significant role in both World War I and World War II, eventually becoming the leader of the free world during the Cold War with the Soviet Union. The United States experienced social and cultural shifts during this time, including the Civil Rights Movement and the rise of consumerism.\\
Today, the United States remains a powerful nation, with a diverse population and a complex history that continues to shape its identity.

\end{prompt}

\begin{prompt}[title={Response \thetcbcounter: Response from Zero-shot CoT in creative writing tasks}, label=creative_cot_response]

\{\textcolor{blue}{\textbf{instruction}}\}\\
Describe the history of the United States.

\{\textcolor{blue}{\textbf{response}} from Zero-shot CoT\}\\
**Pre-Columbian Era**\\
The history of the United States begins long before the arrival of European explorers in the 15th century. The land was initially inhabited by diverse Native American tribes, such as the Apache, Sioux, Iroquois, and Cherokee. These societies had rich cultures, with their own languages, traditions, and social structures. They lived off the land, flourishing through hunting, fishing, farming, and trade. However, the indigenous populations faced significant challenges, including conflicts, displacement, and diseases brought by the European settlers.\\
**Colonial Period**\\
In the 16th century, European powers began to claim territories in what is now the United States. The Spanish were the first to establish colonies, initially focusing on Central and South America. Later, English, French, and Dutch explorers arrived, establishing their own colonies along the Atlantic coast. These colonies varied in purpose and organization, with the English colonies primarily seeking economic opportunities and religious freedom. Over time, thirteen British colonies emerged, forging their own identities, developing local governments, and experiencing an increase in population due to immigration.\\
**American Revolution and Beyond**\\
Tensions between the British colonies and their colonial masters grew, leading to the American Revolution in 1775. The colonies, unified under the Continental Congress, fought for independence from British rule. The war lasted several years, until the signing of the Treaty of Paris in 1783, which granted the newly formed United States of America independence. After achieving independence, the founding fathers convened to draft the Constitution, establishing the framework for the new nation's government.\\
This is just a brief overview, and the history of the United States is vast and complex. However, these steps provide a simplified understanding of how the nation evolved from Native American civilizations to a group of British colonies, to an independent country.

\end{prompt}

\begin{prompt}[title={Response \thetcbcounter: Response from RAG(1-shot) in creative writing tasks}, label=creative_rag1_response]

\{\textcolor{blue}{\textbf{instruction}}\}\\
Describe the history of the United States.

\{\textcolor{blue}{\textbf{response}} from RAG-1\}\\
The history of the United States is a complex and rich tapestry that spans over four centuries. The story begins with the arrival of European explorers and settlers in the 16th century. In 1492, Christopher Columbus reached the Americas, which led to subsequent expeditions and colonization by the Spanish, French, Dutch, and English. The first successful English settlement was established at Jamestown, Virginia in 1607.\\
Over the next few centuries, the colonies grew and developed their own unique cultures, economies, and systems of government. Tensions rose between the American colonies and the British Crown, leading to the American Revolutionary War in 1775. The colonies declared their independence in 1776 and formed the United States of America, which became a republic based on democratic ideals and principles.\\
The young nation faced numerous challenges in its early years, including establishing a central government, expanding westward, and dealing with the issue of slavery. The 19th century saw rapid westward expansion, the Industrial Revolution, and the Civil War. The United States emerged from the Civil War as a more unified and powerful nation, with the abolition of slavery and the preservation of the Union. In the 20th century, the United States became a global superpower, playing a crucial role in both World Wars and the Cold War. It witnessed significant social movements for civil rights and equality, as well as advancements in technology and innovation.\\
Today, the United States is a diverse and influential country with a strong economy, global military presence, and a complex political landscape. Its history continues to shape its present and future, as the nation grapples with ongoing social, economic, and political challenges.

\end{prompt}

\begin{prompt}[title={Response \thetcbcounter: Response from RAG(5-shot) in creative writing tasks}, label=creative_rag5_response]

\{\textcolor{blue}{\textbf{instruction}}\}\\
Describe the history of the United States.

\{\textcolor{blue}{\textbf{response}} from RAG-5\}\\
The history of the United States dates back to the colonization of the Americas by European explorers in the 16th century. The first permanent English settlement was established in Jamestown, Virginia in 1607. This marked the beginning of British colonization in North America.\\
Over the next century, more British colonies were established along the eastern seaboard, including Massachusetts, New York, Pennsylvania, and Georgia. These colonies grew and developed their own distinct cultures and economies.\\
Tensions between the colonies and the British government grew in the mid-18th century, leading to the American Revolutionary War in 1775. The colonists, led by figures such as George Washington, Thomas Jefferson, and Benjamin Franklin, fought for their independence from British rule.\\
In 1776, the Second Continental Congress adopted the Declaration of Independence, which declared the thirteen American colonies to be independent states. This marked the birth of the United States of America.\\
Following the Revolutionary War, the Founding Fathers drafted the Constitution of the United States, which established a federal government and outlined the rights and freedoms of American citizens. The Constitution was ratified in 1788 and is still the supreme law of the land.\\
Since its founding, the United States has experienced significant growth and development. It expanded westward through the Louisiana Purchase and the annexation of Texas, and it eventually stretched from coast to coast. The country also faced significant challenges, such as the Civil War, the Great Depression, and the Civil Rights Movement.\\
Today, the United States is a global superpower with a diverse population and a strong economy. It is known for its democratic system of government, individual freedoms, and cultural influence.

\end{prompt}

\begin{prompt}[title={Response \thetcbcounter: Response from RAT in creative writing tasks}, label=creative_rat_response]

\{\textcolor{blue}{\textbf{instruction}}\}\\
Describe the history of the United States.

\{\textcolor{blue}{\textbf{response}} from RAT\}\\
The history of the United States is characterized by several significant phases. The pre-colonial period witnessed diverse Native American populations with unique cultures inhabiting the land. This era transitioned into colonialism through European exploration and settlement by powers such as the Spanish, French, Dutch, and British, leading to the establishment of colonies along the East Coast with distinct economic and political frameworks.\\
A pivotal moment in American history was the American Revolution, beginning in 1775, as the colonies sought independence from British rule. The Declaration of Independence in 1776 marked the nation's founding on principles of liberty and democracy, establishing autonomy from Great Britain. This revolutionary period not only birthed a republic based on the will of the people but also forged American national identity, fostering a collective ethos entrenched in history and culture, inspiring revolutions worldwide and aiding in the establishment of republican governments elsewhere.\\
The United States stands out as the first modern nation to achieve independence through a national war of liberation, articulating its motives and aspirations in a declaration of independence. This event had global reverberations, spreading fundamental values like liberty, equality, natural and civil rights, and civic duty. The formative years were shaped by the drafting of the Constitution in 1787, defining the federal government's framework. The subsequent ratification of the Bill of Rights further cemented individual freedoms and outlined governance principles that have endured across the nation's history, ensuring a system of checks and balances that protect citizens' rights.\\
The 19th century in the United States witnessed significant developments, including westward expansion, industrialization, and conflicts like the Civil War (1861-1865), which decisively addressed issues of slavery and reinforced federal authority over the states. The late 19th and early 20th centuries were marked by rapid industrial growth, extensive immigration, urbanization, and overseas expansion, including involvement in conflicts such as the Spanish-American War and World War I. The 20th century saw the United States emerge as a global superpower after World War II, engaging in the Cold War with the Soviet Union and participating in conflicts like the Korean and Vietnam Wars, significantly impacting the nation's foreign policy and global standing.\\
Recent history includes social movements for civil rights, gender equality, and environmental protection, as well as economic shifts, advancements in technology, and challenges in areas such as healthcare and international relations.

\end{prompt}

%% file: tables/contamination.tex
\begin{table}[H]
\centering
\captionsetup{justification=centering}
\caption{Benchmark contamination study based on CodeLlama-7b-Python models. We find using the code corpus finetuning the CodeLlama models directly will damage model performance on the code generation benchmark, while using RAT for in-context generation yields better performance.}
\label{tab:contamination}
\resizebox{0.5\linewidth}{!}{
\renewcommand\arraystretch{1.1}
\begin{tabular}{@{}lcclcc@{}}
\toprule
\multirow{2}{*}{Method} & \multicolumn{2}{c}{HumanEval} &  & \multicolumn{2}{c}{HumanEval+} \\ \cmidrule(lr){2-3} \cmidrule(l){5-6} 
                        & pass@1        & pass@5        &  & pass@1         & pass@5        \\ \midrule
DIRECT                  & 40.85\%       & 53.65\%       &  & 37.43\%        & 48.78\%       \\
FINETUNE                & 29.02\%       & 40.24\%       &  & 26.34\%        & 35.98\%       \\
RAT                     & 45.73\%       & 59.75\%       &  & 43.29\%        & 53.66\%       \\ \bottomrule
\end{tabular}}
\end{table}

%% file: tables/minecraft_links.tex
\begin{table}[]
\centering
\captionsetup{justification=centering}
\caption{Page link of retrieved text in embodied Minecraft planning tasks.}
\label{tab:embodied_planning_link}
\resizebox{0.95\linewidth}{!}{
\renewcommand\arraystretch{1.1}
\begin{tabular}{@{}clll@{}}
\toprule
Step & Item              & Recipe                  & Link                                                                   \\ \midrule
1    & 4x Oak Log        & -                       & https://minecraft.fandom.com/wiki/Log                                  \\
2    & 16x Oak Planks    & 4x Oak Log              & https://www.digminecraft.com/basic\_recipes/make\_oak\_wood\_plank.php \\
3    & 4x Stick          & 2x Oak Planks           & https://www.digminecraft.com/basic\_recipes/make\_stick.php            \\
4    & 1x Wooden Pickaxe & 3x Oak Planks, 2 Stick  & https://www.digminecraft.com/tool\_recipes/make\_wooden\_pickaxe.php   \\
5    & 3x Cobblestone    & Wooden Pickaxe          & https://minecraft.fandom.com/wiki/Cobblestone                          \\
6    & 1x Stone Pickaxe  & 3x Cobblestone, 2 Stick & https://www.digminecraft.com/tool\_recipes/make\_stone\_pickaxe.php    \\
7    & 3x Iron Ore       & Stone Pickaxe           & https://minecraft.fandom.com/wiki/Iron\_Ore                            \\
8    & 3x Iron Ingot     & 3x Iron Ore             & https://www.digminecraft.com/basic\_recipes/make\_iron\_ingot.php      \\
9    & 1 Iron Pickaxe    & 3x Iron Ingot, 2x Stick & https://www.digminecraft.com/tool\_recipes/make\_iron\_pickaxe.php     \\
10   & 8x Gold Ore       & Iron Pickaxe            & https://minecraft.fandom.com/wiki/Gold\_Ore                            \\
11   & 8x Gold Ingot     & 8x Gold Ore             & https://www.digminecraft.com/basic\_recipes/make\_gold\_ingot.php      \\
12   & 1x Apple          & -                       & https://minecraft.fandom.com/wiki/Apple                                \\
13   & 1x Golden Apple   & 8x Gold Ingot, 1x Apple & https://www.digminecraft.com/food\_recipes/make\_golden\_apple.php     \\ \bottomrule
\end{tabular}}
\end{table}